\documentclass[letterpaper, 10 pt, conference]{ieeeconf}

\IEEEoverridecommandlockouts  
\newif\ifanonymize
\anonymizefalse

\newcommand{\ourrobot}{\ifanonymize MARBLE\else MARBLE\fi}

\newcommand{\mypara}[1]{\par\addvspace{1mm}\noindent\textbf{#1}}
\providecommand{\tablename}{Table}
\newcommand{\tablecaption}[1]{%
  \refstepcounter{table}%
  \begingroup\footnotesize
  \leftskip=0pt \rightskip=0pt \parfillskip=0pt plus 1fil
  \noindent\MakeUppercase{\tablename}~\thetable: #1\par
  \endgroup
  \vspace{0.5\baselineskip}%
}

\usepackage{cite}
\usepackage{url}
\usepackage{amsmath,amssymb,amsfonts}
\usepackage{algorithmic}
\usepackage{graphicx}
\usepackage{textcomp}
\usepackage{xcolor}
\usepackage{siunitx}
\usepackage{booktabs,array,tabularx}
\usepackage[caption=false,font=footnotesize]{subfig}
\usepackage{adjustbox}
\usepackage{graphicx}
\begin{document}

\sisetup{retain-explicit-plus}

\title{\LARGE \bf
Omnidirectional Amphibious Locomotion via Internal Mass Actuation}

\ifanonymize
  \author{Anonymous Author(s)%
  }
\else
  \author{Niko Weaver$^{*}$, Boxi Xia$^{*}$, Li-Yu Lo$^{*}$, Yuhao Huang,  Boyuan Chen%
  \thanks{*All authors are from Duke University. $^{\dagger}$Equal contribution, co-first authors. This work is supported by DARPA FoundSci program under award HR00112490372, DARPA TIAMAT program under award HR00112490419, ARO under award W911NF2410405, ARL STRONG program under awards W911NF2320182, W911NF2220113, and W911NF242021.}}
\fi

\maketitle
\thispagestyle{empty}
\pagestyle{empty}
\suppressfloats[t] 

\begin{figure}[t]
\centering
\includegraphics[width=\linewidth]{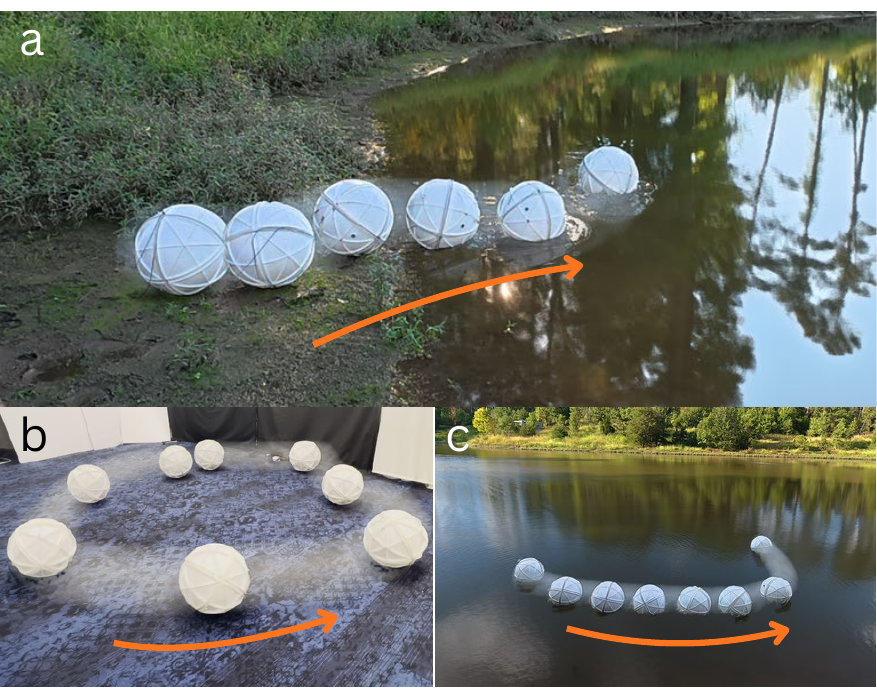}
\vspace{-5px}
\caption{Time-lapse images of \ourrobot{} showing (a) land-to-water entry, (b) terrestrial rolling, and (c) water-surface propulsion. The same internal mass-actuation mechanism produces motion on land and water without mechanical reconfiguration.}
\label{fig:overview}
\end{figure}

\begin{abstract}
Field robots must traverse varied terrain and obstacles while remaining robust to water, debris, vegetation, and physical contact. We present \ourrobot, a fully enclosed omnidirectional amphibious rolling robot driven entirely by internal mass redistribution. Three mutually orthogonal linear sliders shift internal masses to generate body rotation, while an orientation-aware controller maps planar velocity commands into slider positions. A rigid spherical shell encloses all active mechanisms and simultaneously serves as the terrestrial contact surface, buoyant enclosure, and mounting structure for passive fins that enable water-surface propulsion. Rotation of the same shell architecture hence produces rolling on land and surface propulsion in water without mechanical reconfiguration or separate locomotion actuators. The spherical morphology further allows the robot to accommodate changes in body orientation and contact location during direct interactions with terrain and obstacles. We evaluate \ourrobot{} through omnidirectional locomotion characterization, traversal across heterogeneous terrestrial environments, aquatic surface locomotion, land-water transitions, and deliberate obstacle interactions. These experiments demonstrate how a single enclosed mechanical architecture can combine omnidirectional mobility, cross-medium locomotion, and tolerance to environmental contact. \ourrobot{} provides a compact design for field mobility across heterogeneous terrain, obstacles, and land-water transitions. We will open-source all software and hardware design. Our website is 
\url{https://generalroboticslab.com/MARBLE}
\end{abstract}

\section{Introduction}
Field robots operate in environments where terrain, obstacles, and environmental conditions can vary substantially over the course of a mission. Inspection, monitoring, and disaster response may require traversal across uneven ground, vegetation, debris, and aquatic regions~\cite{orekhov2023inspiring,bellicoso2018advances,vogel2014design}. These environments challenge both locomotion and the mechanical structures that produce it. Wheels, limbs, propellers, and other exposed moving components can become obstructed, contaminated, or entangled by vegetation, debris, and water~\cite{briod2014collision,zappetti2022dual,irgens2021experimental}. Robust field mobility depends on the ability to move across heterogeneous environments while tolerating the physical contact and environmental exposure encountered during traversal.

Multimodal robots extend mobility by combining locomotion capabilities for different environments. Air-ground platforms transition between flight and terrestrial motion \cite{kalantari2013hytaq,sihite2023m4,shin2024fast}, while amphibious robots integrate terrestrial and aquatic locomotion~\cite{dudek2007aqua,liang2012amphihex}. Bio-inspired designs further demonstrate how mechanical structures can support multiple locomotion modes. Prior robots have repurposed wings for walking, or coordinated limbs with body motion for both swimming and terrestrial locomotion~\cite{daler2015bioinspired,crespi2013salamandra}. These designs substantially expand the environments accessible to a single robot. However, multiple locomotion modes can introduce additional mechanisms, exposed appendages, or mechanical reconfiguration. Each added mechanism creates another interface between the robot and its surroundings, where water, vegetation, debris, and physical contact can interfere with locomotion.

Spherical morphology offers a way to couple locomotion with mechanical robustness. A spherical body can roll through continuously changing contact locations without relying on a fixed upright orientation, allowing the exterior of the robot to serve directly as its locomotion interface. Spherical, symmetric, and tensegrity robots have used related geometric~\cite{pai1994platonic,nozaki2017spiny,nozaki2018mochibot,vespignani2018superball,surovik2021adaptive} or dynamic properties~\cite{liu2026extreme} for omnidirectional motion, impact tolerance, and locomotion across irregular terrain. Internal actuation can further keep motors and transmissions away from direct environmental contact~\cite{bhattacharya2000spherical,javadi2002introducing,tomik2012design,dejong2017design,ren2023spherical}. Combining these properties creates an opportunity for cross-medium locomotion using a common body and actuation system across different environments.

We present \ourrobot{} (Fig.~\ref{fig:overview}), a fully enclosed omnidirectional amphibious rolling robot driven entirely by internal mass redistribution. Three mutually orthogonal linear sliders shift internal masses to displace the center of mass and generate body rotation. An orientation-aware controller maps planar velocity commands to slider positions as the body continuously reorients. All active components are enclosed within a rigid spherical shell. The shell serves as the rolling surface on land, provides buoyancy in water, and supports passive fins that convert body rotation into water-surface propulsion. The same internal actuation and rotating shell produce both terrestrial and aquatic locomotion without mechanical reconfiguration or separate propulsion actuators. During obstacle interactions, the spherical morphology accommodates changes in body orientation and contact location while the active mechanisms remain enclosed within the shell.

We evaluate \ourrobot{} across terrestrial, aquatic, and obstacle-rich environments. Experiments characterize the directional response of the three-slider actuation system and evaluate locomotion across heterogeneous terrestrial surfaces, water-surface propulsion, land-water transitions, and deliberate obstacle interactions. Our experiments examine how a shared enclosed actuation design can provide omnidirectional mobility across different environments while maintaining mobility during environmental contact. Our main contributions are:
\begin{itemize}
    \item We present a fully enclosed spherical robot design that uses a shared internal mass-actuation mechanism for omnidirectional terrestrial and aquatic locomotion.
    \item We design an orientation-aware controller that coordinates three orthogonal linear mass sliders for omnidirectional rolling under continuous body reorientation.
    \item We experimentally characterize terrestrial rolling, water-surface propulsion, land-to-water transitions, and object interaction through direct shell contact.
\end{itemize}
  
\section{Related Work}
\label{sec:related}

\mypara{Multimodal and Amphibious Locomotion.}
Multimodal robots extend mobility across environments by reusing structures or actuation across locomotion modes. Air-ground robots combine flight with cage-supported rolling~\cite{kalantari2013hytaq,sabet2019rollocopter}, repurpose wings for terrestrial locomotion~\cite{daler2015bioinspired}, or reconfigure appendages for wheeled, legged, and aerial motion~\cite{sihite2023m4}. Land-water robots similarly use shared appendages~\cite{dudek2007aqua}, transformable leg-flipper mechanisms~\cite{liang2012amphihex}, or coordinated spine-limb gaits~\cite{crespi2013salamandra} to move across terrestrial and aquatic environments. These systems demonstrate how shared mechanical structures or mechanical reconfigurations can reduce the hardware dedicated to individual locomotion modes for multimodal robots. Our \ourrobot{} robot extends this idea to a fully enclosed design without reconfiguration. A single internal mass-actuation mechanism rotates the body on both land and water, with the enclosing shell serving as the terrestrial contact surface, buoyant enclosure, and support for passive aquatic propulsion.

\mypara{Spherical Morphology and Terrain Interaction.}
Spherical and highly symmetric robots use distributed contact and body reorientation to maintain mobility across changing orientations and contact conditions. Early multilimbed designs generate rolling through repeated body contacts~\cite{pai1994platonic}, while spherical robots with radial telescopic legs actively vary their shape and contact configuration during locomotion~\cite{nozaki2017spiny,nozaki2018mochibot}. Tensegrity robots use structural compliance to absorb impacts and traverse irregular terrain~\cite{vespignani2018superball,surovik2021adaptive}. More recently, Argus further uses dynamic symmetry and radially distributed actuation to achieve nearly isotropic dynamic capabilities across different body orientations~\cite{liu2026extreme}. These designs show how robot morphology can accommodate changing orientations and contacts during locomotion. \ourrobot{} uses a rigid spherical shell as a continuous external interface for terrain, water, and obstacle contact while keeping its active mechanisms enclosed within the body.
\mypara{Internally Actuated Spherical Robots.}
Internally actuated spherical robots generate locomotion by moving mechanisms inside an outer shell. Prior designs use internal rotors to induce rolling and spinning~\cite{bhattacharya2000spherical}, radial mass redistribution~\cite{javadi2002introducing}, tetrahedrally arranged moving masses~\cite{tomik2012design}, and multiple pendulums~\cite{dejong2017design}. Other designs use dedicated drive and steering mechanisms~\cite{zhan2011design} or omnidirectional wheels acting against the inner surface of the shell~\cite{chen2012design}. Close recent work has extended internal actuation to amphibious locomotion by combining pendulum-driven rolling, rotator-based steering, and passive fins for aquatic propulsion~\cite{chi2021design}. In contrast, \ourrobot{} uses three orthogonal linear mass sliders for orientation-aware omnidirectional locomotion. The same internal actuation drives both terrestrial rolling and water-surface propulsion, avoiding separate steering and propulsion mechanisms while keeping all active locomotion components enclosed.

\section{Mechanical Design and Locomotion}
\label{sec:design}
\begin{figure*}[!t]
    \centering
    \includegraphics[width=0.95\textwidth]{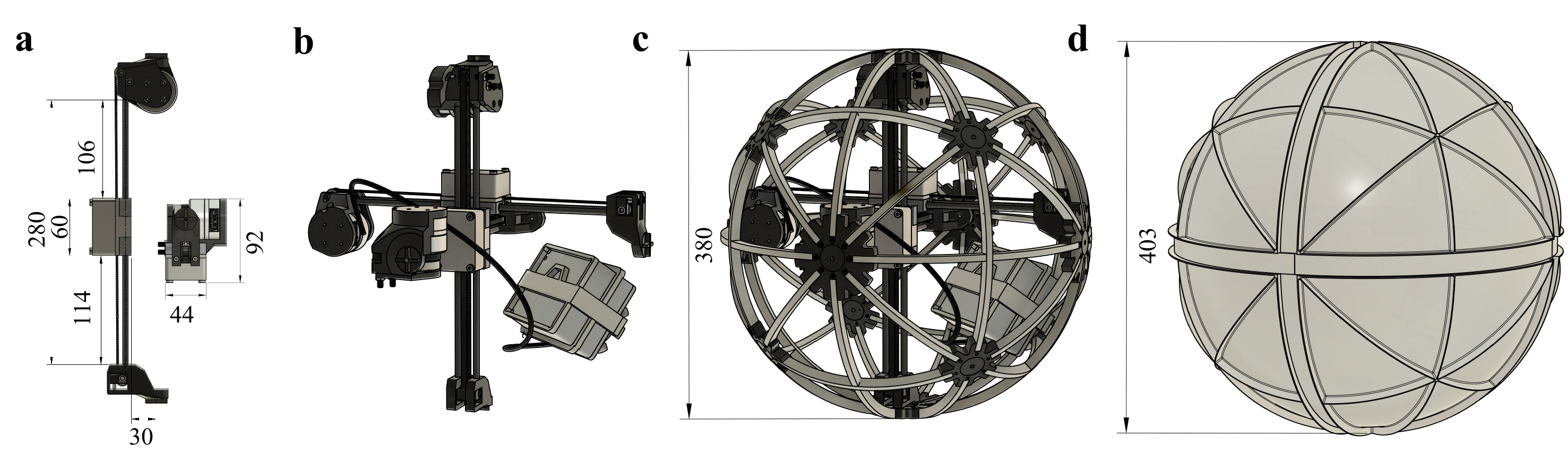}
    \caption{\textbf{Mechanical design of \ourrobot{}.} (a) Linear mass-slider module. (b) Internal configuration with three orthogonal sliders and the electronics capsule. (c) Internal frame without the outer shell. (d) Complete robot with the sealed shell and passive fins.}
    \vspace{-12pt}
    \label{fig:mechanical-design}
\end{figure*}
\subsection{Shell Design}
A rigid spherical shell encloses the structural frame, actuators, and electronics (Fig.~\ref{fig:mechanical-design}(d)) and forms the primary interface between \ourrobot{} and its surroundings. The assembled shell has an outer diameter of \SI{387}{\milli\meter} and a mass of \SI{1.05}{\kilo\gram}. Symmetrically distributed passive fins extend \SI{8}{\milli\meter} beyond the surface. The shell provides the rolling surface for terrestrial locomotion, while the fins generate hydrodynamic forces as the body rotates at the water surface.

The shell is fabricated from polylactic acid (PLA) using fused deposition modeling (FDM). Each hemisphere consists of four identical printed segments. A radial silicone O-ring seals the interface between the two hemispheres, which are secured together by two cast silicone retaining bands with Shore A hardness 10 (Smooth-On Dragon Skin 10). Three plastic bags surrounding the internal components provide secondary protection against water ingress.

The sealed shell also provides buoyancy for water-surface locomotion. The shell encloses a nominal internal volume of \SI{0.030}{\cubic\meter}, with the equilibrium draft determined by the balance between the total robot weight and the buoyant force generated by the displaced external volume. Because the permeability of FDM-fabricated parts depends on print and assembly parameters, we characterize the sealing performance experimentally to find the submerged depth of \ourrobot{}. Our design allows the same exterior structure to serve as the terrestrial rolling surface, buoyant enclosure, and passive aquatic propulsion interface.

\subsection{Structural Frame}
The internal frame holds the three slider modules and onboard electronics (Fig.~\ref{fig:mechanical-design}(b, c)). It consists of 48 beams and 14 hubs, with three beam variants connecting six primary hubs and eight secondary hubs through snap-fit joints. The snap-fit construction eliminates 96 fasteners compared with an assembly using two fasteners per beam. Secondary hubs and connecting beams fill the gaps between the primary structural members and produce a more continuous rolling external profile. This frame provides direct access to the actuators and electronics during development and testing. All field experiments use the complete sealed shell.

\subsection{Mass-Slider Actuation}
Locomotion is generated by three identical linear mass sliders aligned with mutually orthogonal body axes (Fig.~\ref{fig:mechanical-design}(a,b)). Each module translates a \SI{700}{\gram} mass over a total stroke of \SI{220}{\milli\meter}. A GT2 timing belt drives the mass along carbon-fiber guide rods with igus iglidur linear bearings. The three rails are offset by \SI{30}{\milli\meter} to provide clearance between the moving masses.

Each slider is driven by a CubeMars GL40 II motor under joint-level proportional-derivative (PD) position control at \SI{100}{\hertz}. An additional support bearing reduces cantilever loading on the drive assembly under belt tension. By coordinating the slider positions, \ourrobot{} can shift its center of mass along arbitrary directions within the reachable mass-redistribution workspace. The nominal center of mass depends on the rail offsets and the distribution of the fixed components when all three sliders are centered. The sliders operate at a nominal 16 volts from a 4S lithium polymer battery. The motors draw up to \SI{1.88}{\ampere}, have a peak torque of \SI{0.68}{\newton\meter} and peak speed of \SI{145.4}{\radian\per\second}.

\begin{figure}[!t]
\centering
\includegraphics[width=\linewidth]{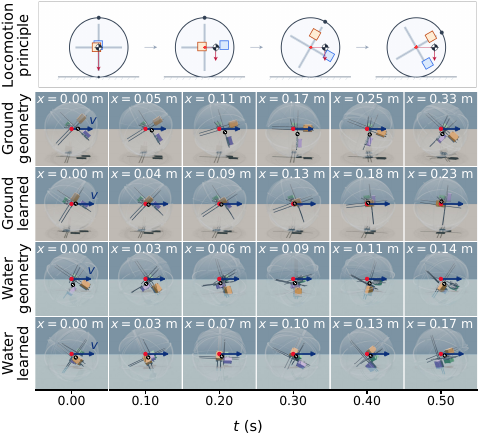}
\vspace{-20pt}
\caption{\textbf{Locomotion principle of \ourrobot{}.} Top: displacing the internal slider masses offsets the center of gravity from the shell center, and the resulting moment about the ground contact rolls the shell. Below: simulation of ground and water-surface locomotion at \SI{0.5}{\meter\per\second} using geometric and learned controllers, sampled at the same instants. The transparent shell shows the three sliding masses, colored one per rail. The quartered circle denotes their center of gravity, the red dot the shell center, and the blue arrow the heading direction.}
\vspace{-12pt}
\label{fig:locomotion}
\end{figure}

\subsection{Locomotion Principle}
Let the body frame $B$ be fixed to the shell with its origin at the geometric center. The center of mass expressed in $B$ is
\begin{align}
\mathbf{r}_G^B
&=
\mathbf{r}_{G,0}^B
+ \frac{1}{M}\sum_{i=1}^{3}
m_i q_i \mathbf{e}_i^B,
\label{eq:center_of_mass}
\end{align}
where $\mathbf{r}_{G,0}^B\in\mathbb{R}^3$ is the nominal center of mass with the sliders centered and $M$ is the total robot mass. For each slider $i\in\{1,2,3\}$, $m_i\in\mathbb{R}_{>0}$ denotes the moving mass, $q_i\in\mathbb{R}$ its displacement from the centered position, and $\mathbf{e}_i^B\in\mathbb{R}^3$ is the corresponding unit rail direction and translation axis. The three axes are mutually orthogonal.

On land, mass redistribution creates a horizontal offset between the center of mass and the ground contact. Gravity then produces a moment that rotates the body. For a horizontal lever arm $d$, the gravitational moment has magnitude $\tau_g=Mgd$. With sufficient tangential traction, the shell rolls without slip. As the body rotates, the controller continuously updates the slider positions to maintain the center-of-mass bias toward the commanded direction. This simplified model captures the primary gravitational mechanism and neglects transient inertial effects, slip, and discrete contacts introduced by the fins.

On the water's surface, the same mass redistribution offsets the center of mass from the line of action of buoyancy and generates body rotation. The submerged fins experience hydrodynamic forces as the shell rotates. Fins above the waterline encounter substantially lower resistance, producing an asymmetric interaction over the rotating shell and net surface propulsion~\cite{chi2021design}. Aquatic motion uses the same three active sliders as terrestrial locomotion. The change in locomotion mode arises from the interaction between the rotating shell and the surrounding medium. The resulting thrust depends on the immersion depth, fin geometry, and body rotation rate and is characterized experimentally.

\subsection{Sensors and Computing}

Onboard computation is provided by an Orange Pi Zero 3W. The computer interfaces over USB with a TM171 nine-axis inertial measurement unit (IMU) and a Seeed Studio XIAO nRF52840 Sense Plus microcontroller. The microcontroller communicates with the three motor drives through an MCP2515-based CAN interface. The IMU estimates body orientation for transforming motion commands into the rotating body frame. For experimental evaluation, robot position and velocity are measured externally from video. 

\section{Control}
\label{sec:control}

We develop two types of controllers with a common low-level actuation interface. An orientation-aware geometric controller directly maps planar motion commands to slider positions using the measured body orientation. A learned controller predicts the same three slider position targets from onboard observations. Both controllers use the joint-level position control. Figure~\ref{fig:locomotion} shows the resulting mass redistribution for both controllers during simulated terrestrial and aquatic locomotion.

\subsection{Coordinate Frames and Command Mapping}
We define a magnetic north--west--up world frame $W$, a body frame $B$ centered on the robot with axes aligned with the three sliders, an IMU frame $S$, and an operator command frame $C$. The rotation matrix $\mathbf{R}_{AB}\in\operatorname{SO}(3)\subset\mathbb{R}^{3\times3}$ maps vectors expressed in frame $B$ into frame $A$. The body orientation is computed using the measured IMU orientation and the fixed IMU mounting transform,
\begin{equation}
  \begin{aligned}
    \mathbf{R}_{WB} &= \mathbf{R}_{WS} \mathbf{R}_{BS}^{\top}, \\
    \mathbf{R}_{WS} &= \operatorname{diag}(1,-1,-1)
    \mathbf{R}_{W_{\mathrm{NED}}S},
  \end{aligned}
  \label{eq:imu-mount}
\end{equation}
where $W_{\mathrm{NED}}$ denotes the IMU's magnetic north--east--down reference frame and $\operatorname{diag}(1,-1,-1)\in\operatorname{SO}(3)$ converts it to the world-frame convention.

The command frame $C$ provides a fixed heading for operator input. Its orientation differs from $W$ by a constant yaw offset $\psi_0\in[-\pi,\pi)$:
\begin{equation}
  \mathbf{R}_{WC} =
  \begin{bmatrix}
    \cos\psi_0 & -\sin\psi_0 & 0 \\
    \sin\psi_0 & \phantom{-}\cos\psi_0 & 0 \\
    0 & 0 & 1
  \end{bmatrix}
  \in\operatorname{SO}(3).
  \label{eq:command-frame}
\end{equation}
The operator sets $\psi_0$ once per run.

For the geometric controller, the planar velocity input $\mathbf{v}^{c}_{xy}=[v^{c}_{x},v^{c}_{y}]^{\top}\in\mathbb{R}^{2}$ specifies the desired direction of travel. A nonzero input defines a virtual target at a fixed distance $\delta$ from the robot center,
\begin{equation}
  \begin{aligned}
    \mathbf{p}^{C}
    &=
    \begin{cases}
      \displaystyle
      \frac{\delta}{\lVert\mathbf{v}^{c}_{xy}\rVert}
      \begin{bmatrix} v^{c}_{x} \\ v^{c}_{y} \\ 0 \end{bmatrix},
      & \lVert\mathbf{v}^{c}_{xy}\rVert > 0, \\[6pt]
      \mathbf{0}_3,
      & \lVert\mathbf{v}^{c}_{xy}\rVert = 0,
    \end{cases} \\
    \delta &= \SI{1.694}{\meter}.
  \end{aligned}
  \label{eq:target-world}
\end{equation}
The normalization removes the command magnitude, so this controller uses a binary stop or move command with a continuous planar direction. The target expressed in the rotating body frame is
\begin{equation}
  \mathbf{p}^{B} = \mathbf{R}_{WB}^{\top} \mathbf{R}_{WC} \mathbf{p}^{C}.
  \label{eq:target-base}
\end{equation}

\subsection{Geometric Controller}

Each slider $i$ moves along a line in the body frame,
\begin{equation}
    \boldsymbol{\ell}_i(q_i)
      = \mathbf{a}_i^B + q_i\mathbf{e}_i^B,
      \qquad i\in\{1,2,3\},
  \label{eq:slider-line}
\end{equation}
where $\mathbf{a}_i^B$ is the mass position at $q_i=0$ and $\mathbf{e}_i^B$ is the corresponding unit rail direction. The rail axes satisfy
\begin{equation}
  (\mathbf{e}_i^B)^\top\mathbf{e}_j^B = 0,
  \qquad i\neq j.
\end{equation}

We project the virtual target onto each slider axis to obtain the desired mass displacement,
\begin{equation}
  q_i^{\star}
    = (\mathbf{e}_i^B)^{\top}
       \bigl(\mathbf{p}^{B}-\mathbf{a}_i^B\bigr).
  \label{eq:projection}
\end{equation}
This projection selects the point along each rail closest to the common virtual target. As the robot rotates, $\mathbf{p}^{B}$ changes continuously and redistributes the three masses to maintain the commanded world-relative direction. The commanded displacement is clipped to the calibrated travel range,
\begin{equation}
    q_i^{\mathrm{cmd}} = \operatorname{clip}\bigl(q_i^{\star},q_{i\min},q_{i\max}\bigr).
  \label{eq:clip}
\end{equation}
Here, $\operatorname{clip}(x,a,b)=\min\{\max\{x,a\},b\}$ for $x,a,b\in\mathbb{R}$ with $a\le b$. A \SI{220}{\milli\meter} stroke provides nominal limits of \SI{+114}{\milli\meter}, \SI{-106}{\milli\meter}, before applying a \SI{\pm 10}{\milli\meter} safety margin. Zero motion command returns the masses to their centered configuration.

The slider displacement is converted to a motor angle setpoint through the belt transmission,
\begin{equation}
    \theta_i = \theta_{i,0} + \sigma_i \frac{2\pi}{N_p p_{\mathrm{GT2}}}q_i^{\mathrm{cmd}},
  \label{eq:motor}
\end{equation}
where $\theta_{i,0}\in\mathbb{R}$ is the encoder angle calibrated at the rail center and $\sigma_i\in\{-1,+1\}$ accounts for transmission and encoder polarity. Each transmission uses a pulley with $N_p=50\in\mathbb{Z}_{>0}$ teeth and a GT2 belt with pitch $p_{\mathrm{GT2}}=\SI{2}{\milli\meter}$. The displacement $q_i^{\mathrm{cmd}}$ and belt pitch $p_{\mathrm{GT2}}$ are expressed in the same length units. The final angle setpoints are transmitted to the three motor drives over CAN.

Overall, the geometric controller computes slider position setpoints by projecting a commanded virtual target onto each rail (Fig.~\ref{fig:locomotion}). At each update, the measured body orientation is used to transform the target into body coordinates. The projected displacements are then clipped to the sliders' travel limits and converted to motor commands, redistributing the internal masses to create a rolling motion.

\subsection{Learned Controller}\label{SCM}

We additionally train learned controllers for terrestrial and aquatic locomotion with reinforcement learning. Both policies share the same hardware interface of the geometric controller and output three slider position targets. This common action space allows the controllers to be exchanged at runtime without changes to the low-level motor control.

\mypara{Simulation.} Training uses 4096 parallel MuJoCo~\cite{todorov2012mujoco,zakka2026mjlab} environments on a single GPU, with a \SI{5}{\milli\second} physics timestep and a \SI{50}{\hertz} control rate. The robot mass and inertia are derived from the CAD assembly. For ground contact, we approximate the shell by a sphere of radius \SI{0.202}{\meter}, corresponding to the fin crests on which the robot rolls.

\mypara{Water surface.} Aquatic training requires modeling buoyancy and the interaction between the rotating fins and the water surface. MuJoCo's built-in fluid model assumes a uniform fluid, so it does not provide buoyancy at the water surface. We augment the MuJoCo dynamics by implementing a surface model that applies a force and a torque to the shell at every physics step, comprising buoyancy, the moment produced by the offset between the centers of mass and buoyancy, drag on the submerged shell, and thrust from the fins. The water is a flat surface at a fixed height $z_w$, with no waves or current. The sealed shell displaces water as a sphere of radius $R=\SI{0.194}{\meter}$, smaller than the \SI{0.202}{\meter} sphere used for ground contact, because water passes between the protruding fins instead of being displaced by them. The fins contribute only through the thrust term.

The buoyant force acts upward at the centroid of the submerged spherical cap, with magnitude $\rho_w g V_{\mathrm{sub}}$, where $\rho_w$ is the water density, $g$ the gravitational acceleration, and $V_{\mathrm{sub}}=\pi h^{2}(3R-h)/3$ the cap volume at submergence depth $h$. The sliders displace the center of mass horizontally while the center of buoyancy stays on the vertical line through the cap centroid, and this offset produces the moment that drives shell rotation in water. The submerged shell also experiences a quadratic drag force that opposes its velocity and scales with the submerged frontal area, for which we use the smooth-sphere drag coefficient of 0.47.

During water locomotion the robot is only partially submerged. Thrust comes from the fins below the waterline. We compute this thrust element-wise, discretizing the fins into short segments and treating each as a slender rod that resists flow across its axis $\hat{\mathbf{t}}_i$ and slips along it. The segments are arranged on 13 rings distributed over the shell, positioned so that every rotation axis has rings that load, since a ring rotating about its own normal slides along itself and does not generate force. These rings do not carry mass, inertia, or collision geometry and only mark where the drag law is evaluated, so the simulated body remains the same as derived from CAD. Each ring is a tube seated on the outside of the shell, with radius $r_t=\SI{8}{\milli\meter}$ equal to the fin height. Dividing each ring into 24 segments resolves the waterline to \SI{15}{\degree} of arc. Only the component of the segment velocity $\mathbf{v}_i$ normal to the segment axis contributes, and expressing all quantities in the world frame,
\begin{equation}
  \begin{aligned}
    \mathbf{F}_i &= -g^{\mathrm{wet}}_i\,g^{\mathrm{vent}}_i\,
      \tfrac{1}{2}\rho_w\,C_r A_s
      \lVert\mathbf{v}_{\perp i}\rVert\,\mathbf{v}_{\perp i}, \\[2pt]
    \mathbf{v}_{\perp i} &= \mathbf{v}_i
      - (\mathbf{v}_i\!\cdot\!\hat{\mathbf{t}}_i)\hat{\mathbf{t}}_i.
  \end{aligned}
  \label{eq:ring-drag}
\end{equation}
Here $C_r=1.0$ is the cross-flow drag coefficient of a circular cylinder, and $A_s=2r_t\ell_s$ is the projected area of a segment of arc length $\ell_s$. The segment velocity $\mathbf{v}_i$ follows from the rigid-body motion of the shell about its center of mass. Each $\mathbf{F}_i$ acts at the center of its segment. Summing the 312 segment contributions gives the thrust and the accompanying torque on the shell.

The wetting factor $g^{\mathrm{wet}}_i$ determines which segments are in the water and is therefore what produces the thrust, since the forces on opposite sides of a fully loaded shell would cancel as it rotates. It rises smoothly from zero to one as a segment submerges over one tube diameter, following $g^{\mathrm{wet}}_i=u_i^{2}(3-2u_i)$ with $u_i=\operatorname{clip}\!\left((z_w-c_{iz})/2r_t,0,1\right)$ for a segment center at height $c_{iz}$. A hard switch at the waterline would instead apply force steps on every revolution, which destabilizes training. The ventilation factor $g^{\mathrm{vent}}_i$ accounts for the air entrained behind a fast-moving segment, which reduces its drag. We model it as $g^{\mathrm{vent}}_i=g_{\min}+(1-g_{\min})/[1+(\mathrm{Fr}_i/\mathrm{Fr}_c)^{2}]$, decaying from one toward $g_{\min}=0.3$ as the segment Froude number $\mathrm{Fr}_i=\lVert\mathbf{v}_{\perp i}\rVert/\sqrt{g\,h}$ passes the onset value $\mathrm{Fr}_c=2$. This number compares the segment speed with the wave speed set by the submergence depth $h$. The factor decays from one toward $g_{\min}=0.3$ above $\mathrm{Fr}_c=2$, the value at which air begins to enter the wake.

The model omits added mass, wave-making drag, and interactions between individual fin wakes. The unmodeled effects can reduce the thrust generated by the robot, so we account for their combined effect using a per-episode scaling of the ring forces (Table~\ref{tab:rl-setup}). The terrestrial and aquatic configurations share the robot model, actuation, observations, and action space. The aquatic configuration adds the water-surface forces and thrust randomization and uses a weaker travel-limit penalty.

\mypara{Policy.} The policy maps a 23-dimensional onboard observation to three slider position targets. Each action is scaled by \SI{0.104}{\meter} about the centered rest pose to cover the usable travel in \eqref{eq:clip}, and tracked by the same joint-level position controller used by the geometric controller. The observation contains body angular velocity, the command in operator frame $C$, slider positions and velocities, the previous action, and the body rotation matrix $\mathbf{R}_{WB}$. The rotation matrix provides a unique, continuous representation across all shell orientations, avoiding Euler-angle singularities and quaternion sign ambiguity. Three consecutive observation frames are stacked and encoded. During training, the critic additionally receives uncorrupted observations and the ground-truth linear velocity. The ground-truth linear velocity is unavailable to the deployed policy and provides privileged information for learning an internal velocity estimate from onboard observation.

\mypara{Training.} Velocity commands are sampled per axis within $\pm 0.7~\mathrm{m/s}$ and resampled every \SIrange{2}{8}{\second}. Ten percent of the sampled commands are zero to include stationary behavior during training. The reward encourages planar velocity tracking while regularizing slider velocity, travel-limit violations, action rate, and action acceleration (Table~\ref{tab:rl-setup}(a)). The tracking kernel widens with command speed to accommodate errors across the range from rest to $0.7~\mathrm{m/s}$. We anneal the action-rate penalty over the first 3000 iterations to avoid suppressing slider motion before a rolling gait develops.

Episodes last \SI{20}{\second}, without failure termination because the spherical body does not have a required upright orientation. Shell contact is part of normal locomotion. Each policy is trained for 15000 iterations using FlashSAC~\cite{kim2026flashsac}, an off-policy soft actor-critic algorithm with a distributional critic. We randomize contact, actuation, mass, and latency parameters during training as summarized in Table~\ref{tab:rl-setup}(b).

\mypara{Deployment.} The learned policy runs onboard at \SI{50}{\hertz} and directly outputs the three slider position targets. Deployment uses the policy trained in simulation without hardware fine-tuning or an additional adaptation phase. The geometric and learned controllers differ only in how the slider targets are generated and share the same sensing, communication, and low-level actuation stack.

\begin{table}[t]
\vspace{7pt}
\centering
\footnotesize
\tablecaption{Reward terms and domain randomization shared by the terrestrial and aquatic policies unless otherwise noted.}
\label{tab:rl-setup}

\begin{tabularx}{\linewidth}{
  @{}l@{\hspace{4pt}}
  >{\raggedright\arraybackslash}X@{\hspace{4pt}}
  r@{}
}
\multicolumn{3}{c}{\textbf{(a) Reward terms}}\\
\toprule
\textbf{Term} & \textbf{Expression $r_i$} & \textbf{$w_i$}\\
\midrule
Velocity tracking & $\exp\bigl(-(\lVert\mathbf{v}^{c}_{xy}-\mathbf{v}_{xy}\rVert^{2}+\lambda_z v_z^{2})/\beta^{2}\bigr)$ & $+2.0$\\
Slider velocity & $\sum_j \dot q_j^{2}$ & $-0.002$\\
Travel limit & $\sum_j\bigl[(q_j-q_j^{\max})_+^{2}+(q_j^{\min}-q_j)_+^{2}\bigr]$ & $-1.0$\\
Action rate & $\lVert\mathbf{a}_t-\mathbf{a}_{t-1}\rVert^{2}$ & $-0.05\!\rightarrow\!-0.2$\\
Action acceleration & $\lVert\mathbf{a}_t-2\mathbf{a}_{t-1}+\mathbf{a}_{t-2}\rVert^{2}$ & $-0.05$\\
\bottomrule
\multicolumn{3}{@{}p{\linewidth}@{}}{\scriptsize Total reward $\sum_i w_i r_i$; $a\!\rightarrow\!b$ is a curriculum. $\mathbf{v}_{xy},v_z$: shell velocity; $\mathbf{v}^{c}_{xy}$: command; $\beta=\max(0.5\lVert\mathbf{v}^{c}_{xy}\rVert,0.3~\mathrm{m/s})$: kernel width; $\lambda_z=0.1$; $q_j$: slider position; $q_j^{\max},q_j^{\min}$: soft travel limits; $\mathbf{a}_t$: action; $(x)_+=\max(x,0)$.}
\end{tabularx}

\vspace{2.5mm}

\begin{tabularx}{\linewidth}{@{}>{\raggedright\arraybackslash}X l c@{}}
\multicolumn{3}{c}{\textbf{(b) Domain randomization and observation noise}}\\
\toprule
\textbf{Parameter} & \textbf{Range} & \textbf{When}\\
\midrule
shell-ground friction & \numrange{0.2}{0.6} (abs) & S\\
Slider PD gains $k_p,k_d$ & $\times\,$\numrange{0.8}{1.2} & R\\
Slider force limit & \SIrange{31.4}{37.7}{\newton} & R\\
Rail Coulomb friction & \SIrange{1}{7}{\newton} & R\\
Slider joint damping & \numrange{0}{0.01} (add) & R\\
Body mass & $\times\,$\numrange{0.85}{1.15} & \SI{9.6}{\second}\\
Shell CoM offset & $\pm\SI{10}{\milli\meter}$ (add) & \SI{9.6}{\second}\\
Command delay (to motors) & \SIrange{20}{80}{\milli\second} & \SI{20}{\milli\second}\\
Sensor delay (to policy) & \SIrange{0}{40}{\milli\second} & \SI{120}{\milli\second}\\
Ring thrust scale\textsuperscript{b} & $\times\,$\numrange{0.6}{1.0} & R\\
\midrule
Angular velocity & $\mathcal{N}(0,0.1)$ & step\\
Rotation matrix & $\mathcal{U}(\pm0.05)$ & step\\
Slider velocity & $\mathcal{U}(\pm0.05)$ & step\\
Slider position & $\mathcal{U}(\pm\SI{5}{\milli\meter})$, bias & step, R\\
\bottomrule
\multicolumn{3}{@{}p{\linewidth}@{}}{\scriptsize \emph{When}: once per environment (S), once per episode (R), or at the stated interval. Rail friction ranges from nearly free up to the \SI{6.95}{\newton} that holds one slider against gravity, and the command delay brackets the \SIrange{26}{56}{\milli\second} CAN round trip measured on hardware. \textsuperscript{b}Aquatic only.}
\end{tabularx}
\end{table}

\begin{figure*}[t]
\vspace{7pt}
    \centering
    \begin{minipage}[t]{0.33\textwidth}
        \centering
        \includegraphics[width=\linewidth]{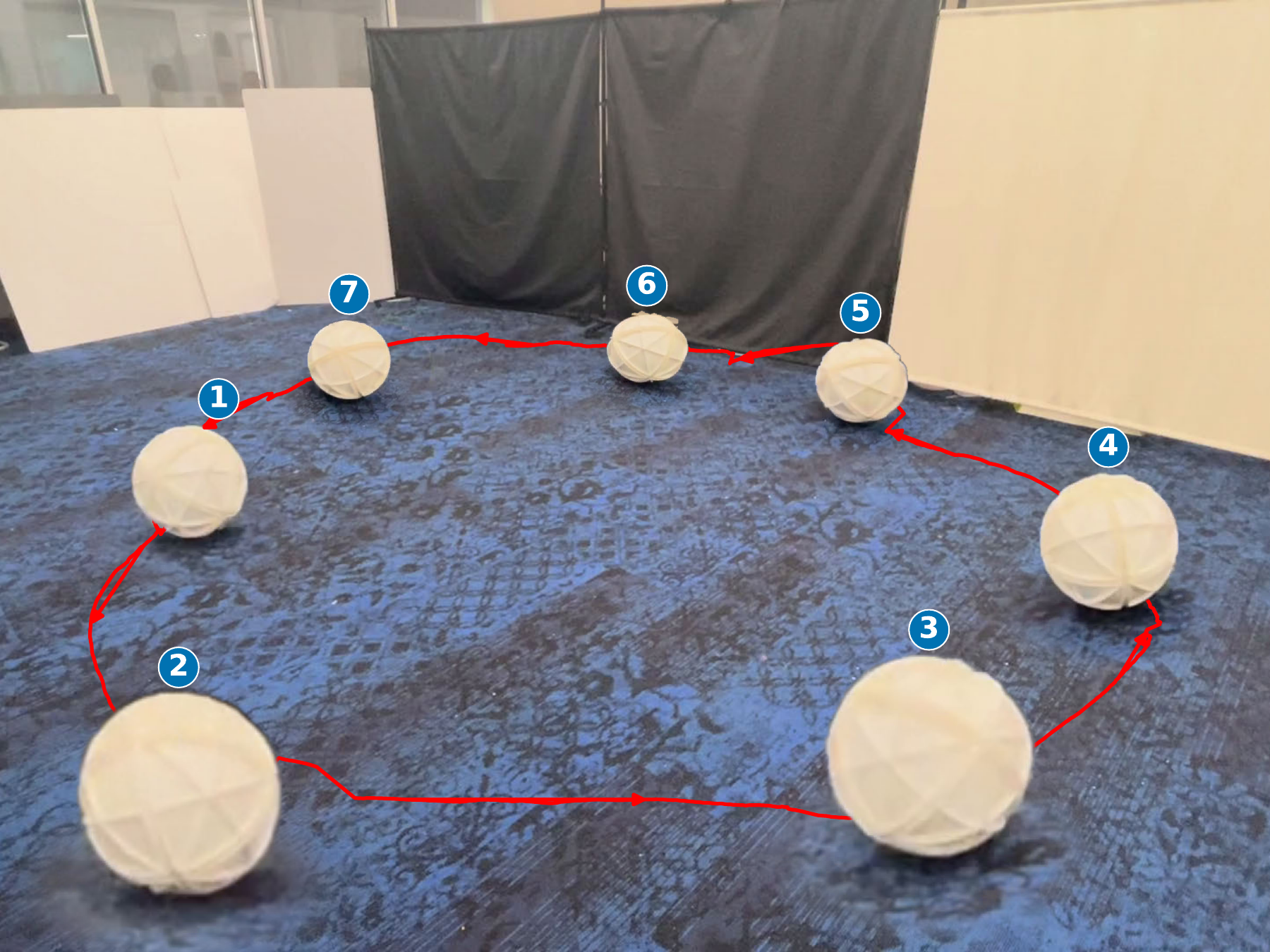}
    \end{minipage}\hfill
    \begin{minipage}[t]{0.33\textwidth}
        \centering
        \includegraphics[width=\linewidth]{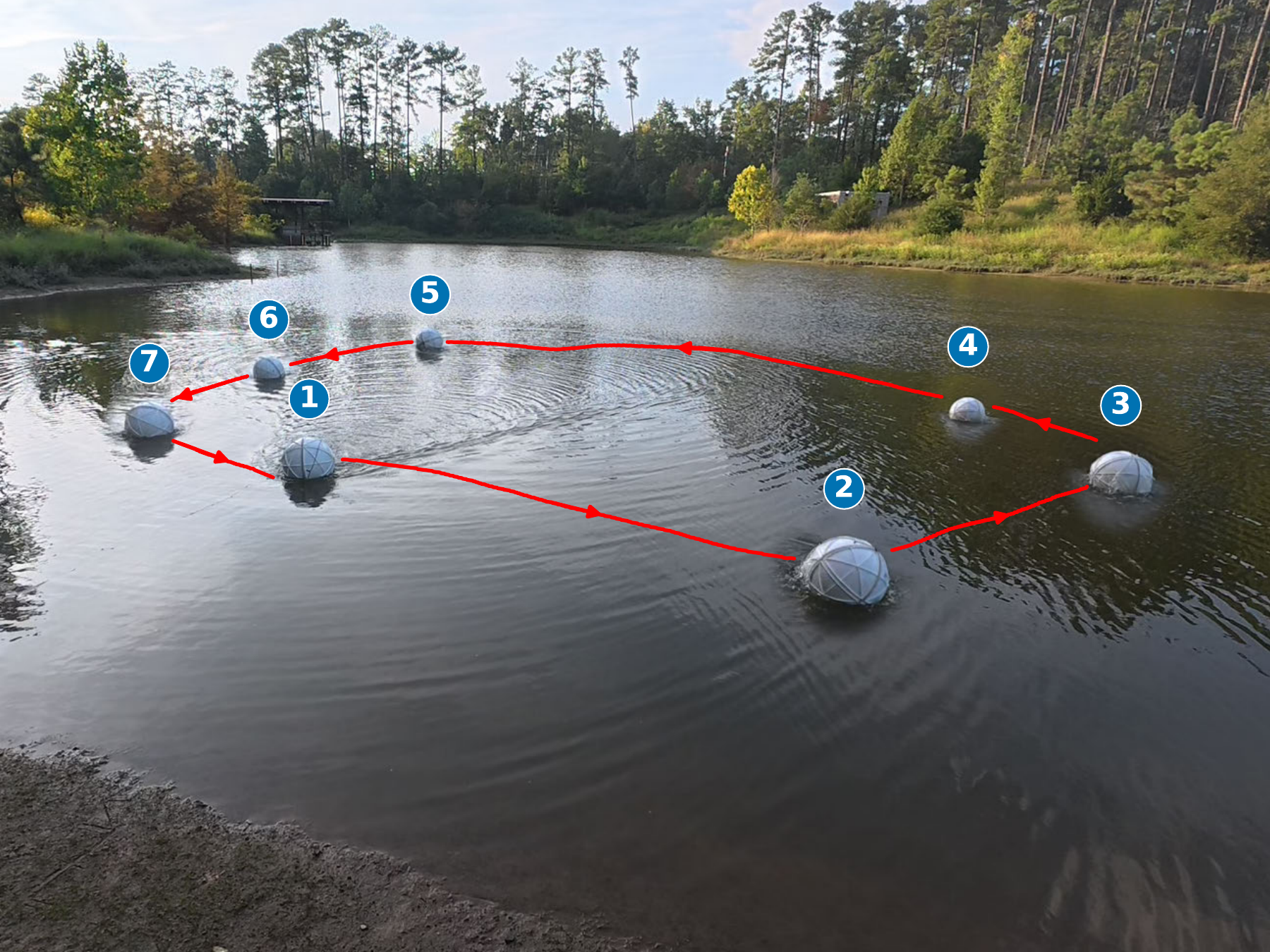}
    \end{minipage}\hfill
    \begin{minipage}[t]{0.33\textwidth}
        \centering
        \includegraphics[width=\linewidth]{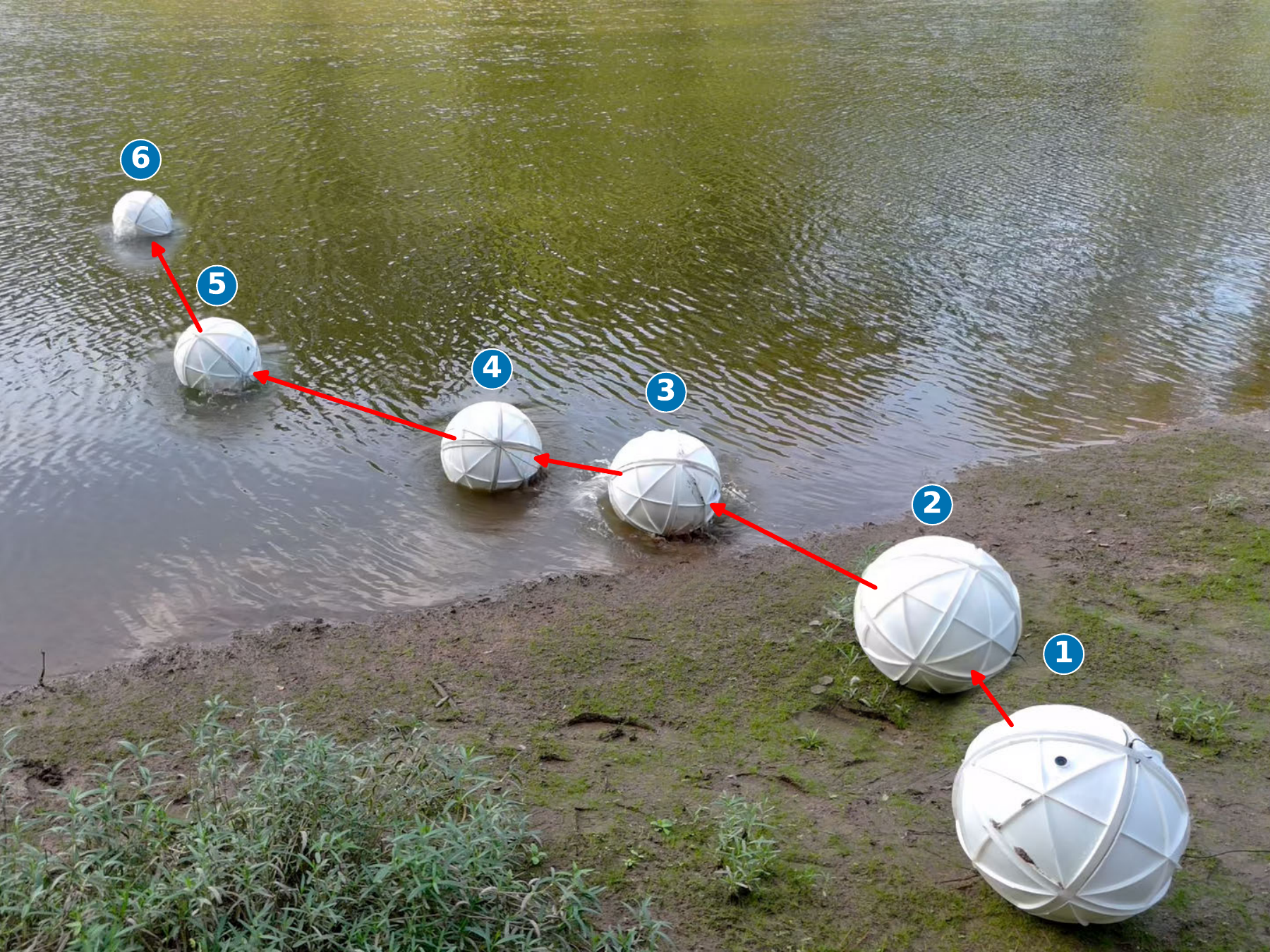}
    \end{minipage}

    \begin{minipage}[t]{0.33\textwidth}
        \centering
        \includegraphics[width=\linewidth]
            {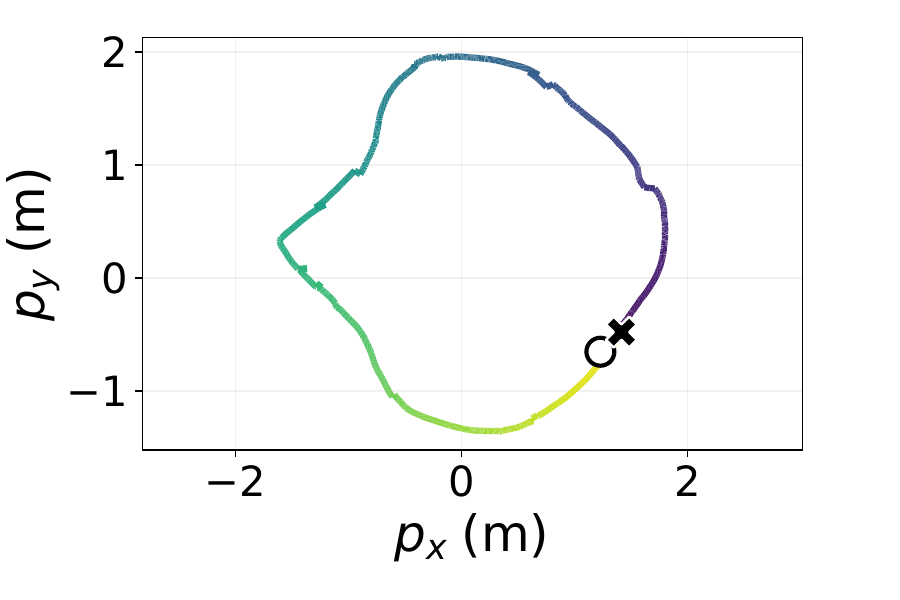}
    \end{minipage}\hfill
    \begin{minipage}[t]{0.33\textwidth}
        \centering
        \includegraphics[width=\linewidth]
            {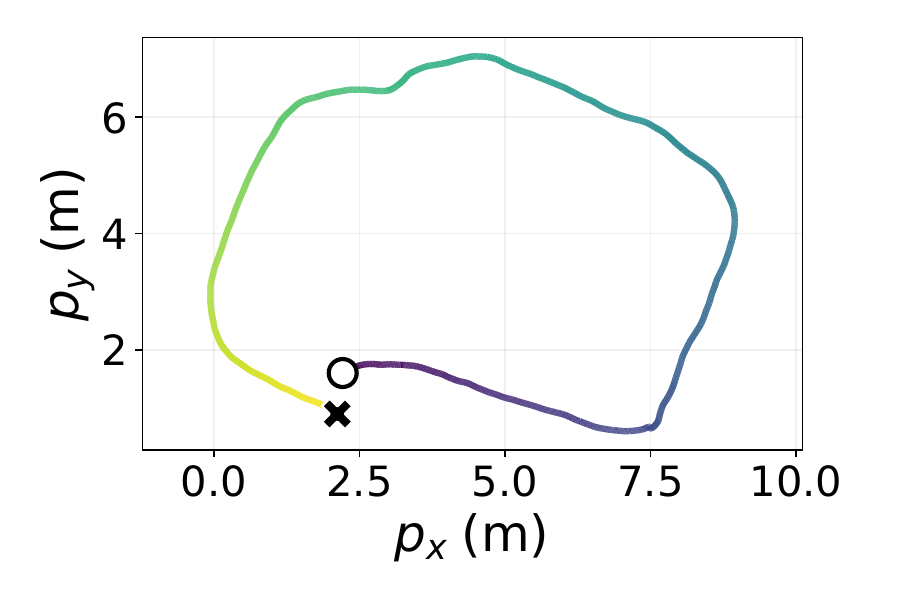}
    \end{minipage}\hfill
    \begin{minipage}[t]{0.33\textwidth}
        \centering
        \includegraphics[width=\linewidth]
            {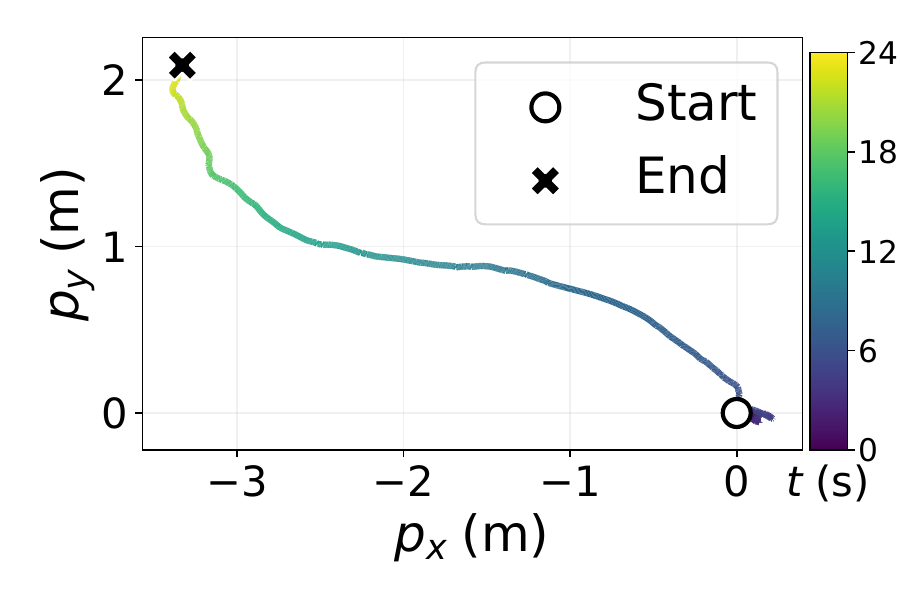}
    \end{minipage}

    \vspace{-10px}
    \caption{Terrestrial, aquatic, and transitional locomotion of \ourrobot{}. 
    Top row: representative time-lapse sequences. 
    Bottom row: corresponding planar position trajectories for 
    (a) terrestrial, (b) aquatic, and (c) transitional locomotion. 
    Trajectory color indicates temporal progression, while the circle and cross mark the start and end positions, respectively.}
    \vspace{-10pt}
    \label{fig:position-trajectories}
\end{figure*}

\newlength{\velocityfigureheight}

\begin{figure}[t]
\centering
\vspace{3pt}
\setlength{\velocityfigureheight}{0.395\linewidth}

\begin{minipage}[t]{0.434\linewidth}
    \vspace{0pt}
    \centering
    \includegraphics[height=\velocityfigureheight]{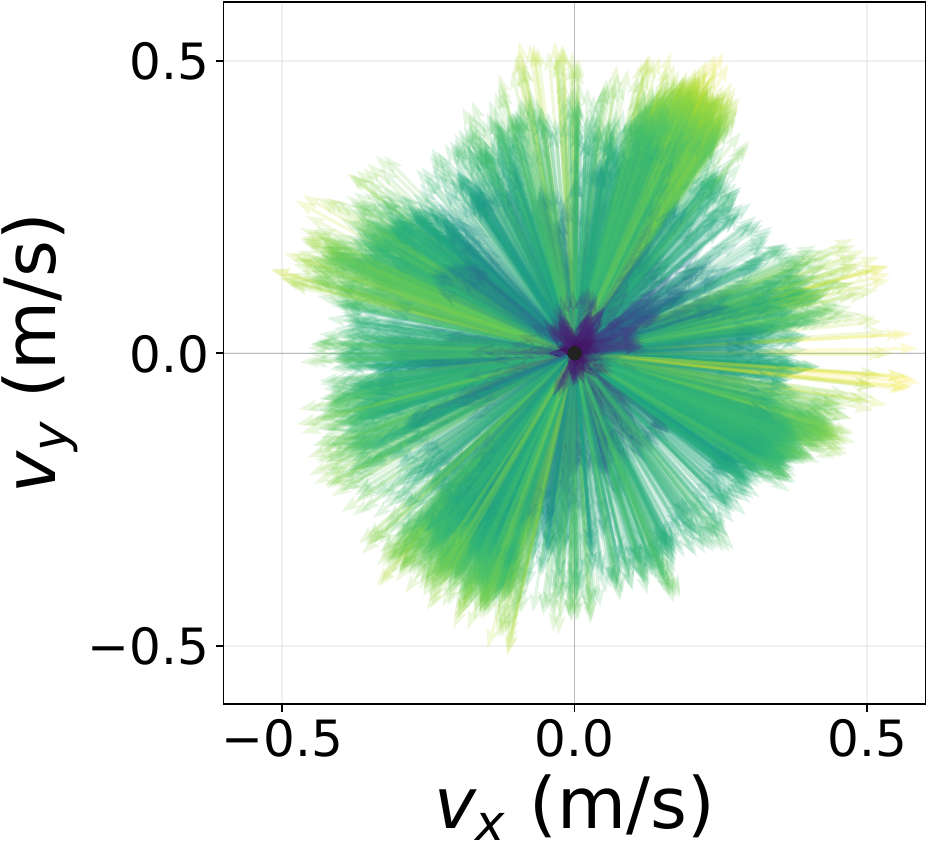}
    \par {\small (a) Learned controller}
\end{minipage}%
\hspace{0.005\linewidth}%
\begin{minipage}[t]{0.556\linewidth}
    \vspace{0pt}
    \centering
    \includegraphics[height=\velocityfigureheight]{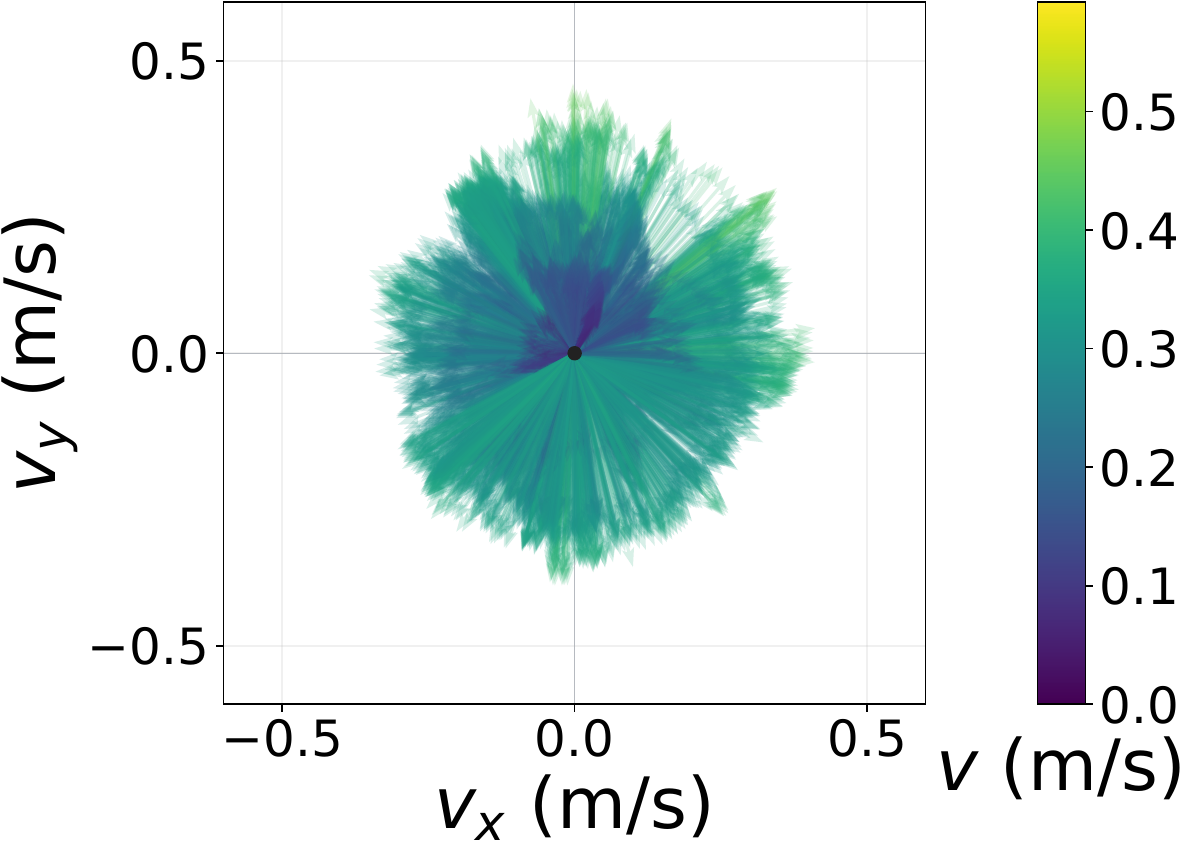}
    \par {\small (b) Geometric controller}
\end{minipage}

\caption{Measured velocity vectors under joystick commands for
(a) the learned controller and (b) the geometric controller.
Velocities are estimated by numerical differentiation of
camera-relative trajectories and plotted from a common origin.
Each panel contains 7,568 samples with identical axis and color scales.}
\vspace{-10pt}
\label{fig:omnidirectional-velocity}
\end{figure}

\begin{figure*}[!t]
\centering
\vspace{3pt}
\includegraphics[width=1.0\textwidth]{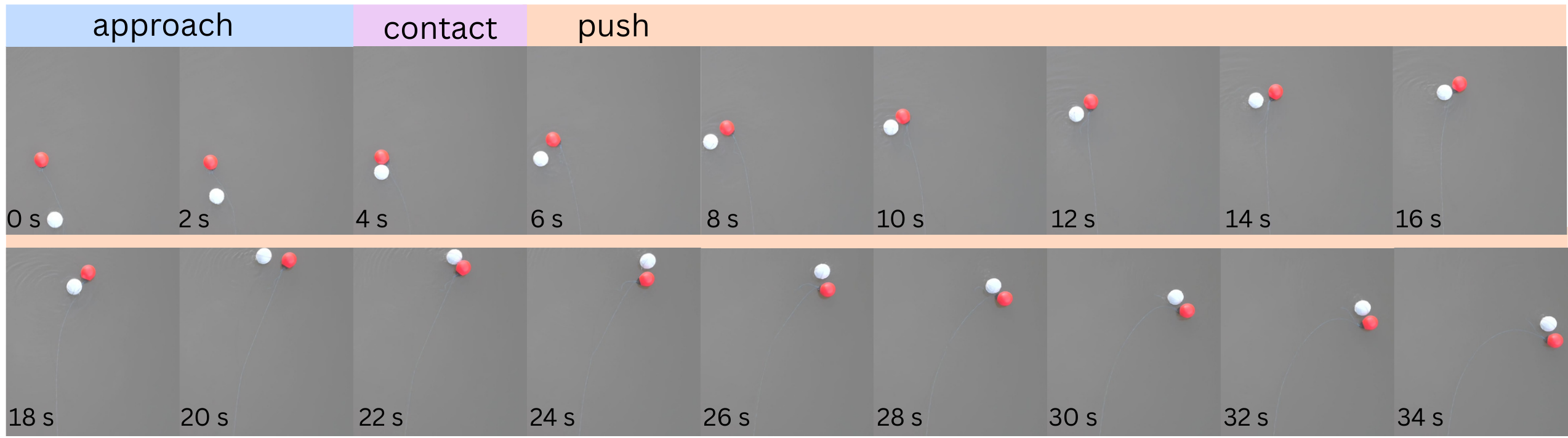}
\vspace{-16pt}
\caption{Representative time-lapse sequence of obstacle interaction showing approach, direct shell contact, and continuous interaction with the buoy (red).}
\label{fig:obstacle-interaction}
\end{figure*}

\section{Experiments}\label{sec:experiments}
\subsection{Terrestrial, Aquatic, and Transitional Locomotion}

We first evaluated whether the shared internal actuation mechanism supported locomotion across terrestrial and aquatic environments. Fig.~\ref{fig:position-trajectories} shows representative time-lapse sequences and reconstructed planar trajectories on land and at the water surface. During terrestrial locomotion, \ourrobot{} traveled \SI{10.60}{\meter} over \SI{14.00}{\second}, reaching a mean speed of $0.745 \pm 0.163~\mathrm{m/s}$ and a maximum speed of $1.025~\mathrm{m/s}$, while following an approximately closed spatial trajectory. On water, \ourrobot{} traveled \SI{5.24}{\meter} over \SI{14.00}{\second}, with a mean speed of $0.375 \pm 0.059~\mathrm{m/s}$ and a maximum speed of $0.488~\mathrm{m/s}$. These results demonstrate sustained terrestrial and aquatic locomotion without introducing a separate aquatic propulsion mechanism.

We next commanded \ourrobot{} to enter the water directly from land. The robot progressed continuously from terrestrial rolling to water-surface propulsion without mechanical reconfiguration or switching propulsion hardware. During the transition, it traveled $5.21~\mathrm{m}$ over $23.93~\mathrm{s}$ with a mean speed of $0.215 \pm 0.146~\mathrm{m/s}$ and a maximum speed of $0.840~\mathrm{m/s}$. The reduced mean speed occurred during shoreline entry, where the robot encountered changing contact and buoyancy conditions. The complete sequence shows continuous progression from terrestrial rolling to sustained aquatic locomotion using the same internal mass actuation.

\subsection{Omnidirectional Velocity Characterization}
We then characterized \ourrobot{}'s planar mobility across different travel directions and compared the geometric controller (GC) with the learned controller (LC). Commands were specified in the operator frame and transformed into the rotating body frame using the measured orientation, as described in Sec.~\ref{sec:control}.

Fig.~\ref{fig:omnidirectional-velocity} presents segmented velocity responses collected across different commanded headings. The distribution of velocity vectors over the full $360^\circ$ range demonstrated that \ourrobot{}'s actuation mechanism can produce planar motion across arbitrary directions without a preferred fixed heading. This capability arose from coordinated mass redistribution along the three mutually orthogonal body axes.

We further compared the two controllers during water-surface locomotion. The learned controller reached a mean speed of $0.348 \pm 0.114~\mathrm{m/s}$ and a maximum speed of $0.592~\mathrm{m/s}$. The geometric controller reached a mean speed of $0.252 \pm 0.088~\mathrm{m/s}$ and a maximum speed of $0.461~\mathrm{m/s}$. The learned controller recording covered $94.10~\mathrm{m}$ over $270.37~\mathrm{s}$, while the geometric controller covered $63.67~\mathrm{m}$ over $252.20~\mathrm{s}$. These results show that both controllers produced omnidirectional aquatic locomotion, with the learned controller reaching higher translational speeds.

\subsection{Object Interaction with Direct Shell Contact}

The spherical morphology of \ourrobot{} is designed to accommodate environmental contact without exposing the active locomotion mechanisms. We evaluated this property by commanding \ourrobot{} to continuously push a buoy and allowing the shell and passive fins to contact the obstacle, as shown in Fig.~\ref{fig:obstacle-interaction}.

During contact, the buoy changed the instantaneous contact location and perturbed the orientation and motion of the robot. \ourrobot{} continued actuating its internal masses throughout the interaction and subsequently disengaged from the buoy and resumed locomotion. The active sliders, motors, and transmissions remain enclosed within the shell throughout the interaction. Fig.~\ref{fig:obstacle-interaction} shows the progression from approach through shell contact and subsequent recovery. These experiments illustrate how the spherical exterior of \ourrobot{} can serve as the contact interface during obstacle interaction while the locomotion mechanism remains internal.
  
\section{Conclusions, limitations, and future work}

We presented \ourrobot{}, a fully enclosed amphibious robot that uses three orthogonal internal mass sliders for omnidirectional locomotion across land and water. The same actuation mechanism rotates the body for terrestrial rolling and drives passive fins for water-surface propulsion, without mechanical reconfiguration or separate active aquatic propulsion. Experiments characterized omnidirectional locomotion with geometric and learned controllers and demonstrated terrestrial rolling, water-surface propulsion, land-water transitions, and continued pushing during obstacle contact. Our results show how a shared internal actuation design can support cross-medium mobility while keeping the active locomotion mechanisms enclosed within the robot body.

The current platform is limited to water-surface locomotion, and its learned aquatic controller relies on an approximate hydrodynamic model. Though our learned controller performs adaptive locomotion, autonomous navigation is not supported. Future work will extend \ourrobot{} toward autonomous field operation through additional onboard sensing and perception. Further development of the aquatic platform will investigate submerged locomotion with improved hydrodynamic modeling. These extensions can expand the same enclosed, mass-actuated design toward adaptive operation in more complex environments.

\bibliographystyle{IEEEtran}
\bibliography{references}

@inproceedings{javadi2002introducing,
  title={Introducing {August}: A Novel Strategy for an Omnidirectional Spherical Rolling Robot},
  author={Javadi A., Amir Homayoun and Mojabi, P.},
  booktitle={Proceedings 2002 IEEE International Conference on Robotics and Automation (ICRA)},
  volume={4},
  pages={3527--3533},
  year={2002},
  publisher={IEEE}
}

@article{dejong2017design,
  title={Design and Analysis of a Four-Pendulum Omnidirectional Spherical Robot},
  author={DeJong, Brian P. and Karadogan, Ernur and Yelamarthi, Kumar and Hasbany, James},
  journal={Journal of Intelligent \& Robotic Systems},
  volume={86},
  number={1},
  pages={3--15},
  year={2017},
  publisher={Springer}
}

@inproceedings{chen2012design,
  title={Design and Implementation of an Omnidirectional Spherical Robot {Omnicron}},
  author={Chen, W.-H. and Chen, C.-P. and Yu, W.-S. and Lin, C.-H. and Lin, P.-C.},
  booktitle={2012 IEEE/ASME International Conference on Advanced Intelligent Mechatronics (AIM)},
  pages={719--724},
  year={2012},
  publisher={IEEE}
}

@inproceedings{zhan2011design,
  title={Design, Analysis and Experiments of an Omni-Directional Spherical Robot},
  author={Zhan, Qiang and Cai, Yao and Yan, Caixia},
  booktitle={2011 IEEE International Conference on Robotics and Automation (ICRA)},
  pages={4921--4926},
  year={2011},
  publisher={IEEE}
}

@article{chi2021design,
  title={Design and Modelling of an Amphibious Spherical Robot Attached with Assistant Fins},
  author={Chi, Xing and Zhan, Qiang},
  journal={Applied Sciences},
  volume={11},
  number={9},
  pages={3739},
  year={2021},
  publisher={MDPI}
}

@article{tomik2012design,
  title={Design, fabrication and control of spherobot: A spherical mobile robot},
  author={Tomik, Filip and Nudehi, Shahin and Flynn, Louis L and Mukherjee, Ranjan},
  journal={Journal of Intelligent \& Robotic Systems},
  volume={67},
  number={2},
  pages={117--131},
  year={2012},
  publisher={Springer}
}

@article{bhattacharya2000spherical,
  title={Spherical rolling robot: A design and motion planning studies},
  author={Bhattacharya, Shourov and Agrawal, Sunil K},
  journal={IEEE Transactions on Robotics and Automation},
  volume={16},
  number={6},
  pages={835--839},
  year={2000},
  publisher={IEEE}
}

@article{liu2026extreme,
  title={Extreme Dynamic Symmetry Enables Omnidirectional and Multifunctional Robots},
  author={Liu, Jiaxun and Xia, Boxi and Chen, Boyuan},
  journal={Science Robotics},
  volume={11},
  number={114},
  pages={eaec1725},
  year={2026},
  publisher={American Association for the Advancement of Science}
}

@article{orekhov2023inspiring,
  title={Inspiring field robotics advances through the design of the darpa subterranean challenge},
  author={Orekhov, Viktor L and Maio, Angela C and Daniel, Roshan P and Chung, Timothy H},
  journal={Field Robotics},
  volume={3},
  pages={560--604},
  year={2023},
  publisher={FRPS}
}

@article{bellicoso2018advances,
  title={Advances in real-world applications for legged robots},
  author={Bellicoso, C Dario and Bjelonic, Marko and Wellhausen, Lorenz and Holtmann, Kai and G{\"u}nther, Fabian and Tranzatto, Marco and Fankhauser, P{\'e}ter and Hutter, Marco},
  journal={Journal of Field Robotics},
  volume={35},
  number={8},
  pages={1311--1326},
  year={2018},
  publisher={Wiley Online Library}
}

@inproceedings{pai1994platonic,
  author = {Pai, Dinesh K. and Barman, R. A. and Ralph, S. K.},
  title = {Platonic Beasts: A New Family of Multilimbed Robots},
  booktitle = {Proceedings of the 1994 IEEE International Conference on Robotics and Automation},
  pages = {1019--1025},
  year = {1994}
}

@inproceedings{nozaki2017spiny,
  author = {Nozaki, Hiroki and Niiyama, Ryuma and Yonezawa, Takuro and Nakazawa, Jin},
  title = {Shape Changing Locomotion by Spiny Multipedal Robot},
  booktitle = {2017 IEEE International Conference on Robotics and Biomimetics},
  pages = {2162--2166},
  year = {2017},
  doi = {10.1109/ROBIO.2017.8324739}
}

@inproceedings{nozaki2018mochibot,
  author = {Nozaki, Hiroki and Kujirai, Yusei and Niiyama, Ryuma and
            Kawahara, Yoshihiro and Yonezawa, Takuro and Nakazawa, Jin},
  title = {Continuous Shape Changing Locomotion of 32-Legged Spherical Robot},
  booktitle = {2018 IEEE/RSJ International Conference on Intelligent Robots and Systems},
  pages = {2721--2726},
  year = {2018},
  doi = {10.1109/IROS.2018.8593791}
}

@inproceedings{vespignani2018superball,
  author = {Vespignani, Massimo and Friesen, Jeffrey M. and
            SunSpiral, Vytas and Bruce, Jonathan},
  title = {Design of {SUPERball v2}, a Compliant Tensegrity Robot for Absorbing Large Impacts},
  booktitle = {2018 IEEE/RSJ International Conference on Intelligent Robots and Systems},
  pages = {2865--2871},
  year = {2018},
  doi = {10.1109/IROS.2018.8594374}
}

@article{surovik2021adaptive,
  author = {Surovik, David and Wang, Kun and Vespignani, Massimo and
            Bruce, Jonathan and Bekris, Kostas E.},
  title = {Adaptive Tensegrity Locomotion: Controlling a Compliant Icosahedron
           with Symmetry-Reduced Reinforcement Learning},
  journal = {The International Journal of Robotics Research},
  volume = {40},
  number = {1},
  pages = {375--396},
  year = {2021},
  doi = {10.1177/0278364919859443}
}

@inproceedings{vogel2014design,
  title={Design of a compliance assisted quadrupedal amphibious robot},
  author={Vogel, Andrew R and Kaipa, Krishnanand N and Krummel, Gregory M and Bruck, Hugh A and Gupta, Satyandra K},
  booktitle={2014 IEEE International Conference on Robotics and Automation (ICRA)},
  pages={2378--2383},
  year={2014},
  organization={IEEE}
}

@article{irgens2021experimental,
  title={Experimental assessment of entanglement for a propeller driven unmanned underwater vehicle},
  author={Irgens, Katherine E and Klamo, Joseph T and Pollman, Anthony G},
  journal={Naval Engineers Journal},
  volume={133},
  number={3},
  pages={103--114},
  year={2021},
  publisher={American Society of Naval Engineers}
}

@article{zappetti2022dual,
  title={Dual stiffness tensegrity platform for resilient robotics},
  author={Zappetti, Davide and Sun, Yi and Gevers, Matthieu and Mintchev, Stefano and Floreano, Dario},
  journal={Advanced Intelligent Systems},
  volume={4},
  number={7},
  pages={2200025},
  year={2022},
  publisher={Wiley Online Library}
}

@article{briod2014collision,
  title={A collision-resilient flying robot},
  author={Briod, Adrien and Kornatowski, Przemyslaw and Zufferey, Jean-Christophe and Floreano, Dario},
  journal={Journal of Field Robotics},
  volume={31},
  number={4},
  pages={496--509},
  year={2014},
  publisher={Wiley Online Library}
}

@article{crespi2013salamandra,
  author = {Crespi, Alessandro and Karakasiliotis, Konstantinos and
            Guignard, Andre and Ijspeert, Auke Jan},
  title = {{Salamandra Robotica II}: An Amphibious Robot to Study
           Salamander-Like Swimming and Walking Gaits},
  journal = {IEEE Transactions on Robotics},
  volume = {29},
  number = {2},
  pages = {308--320},
  year = {2013},
  doi = {10.1109/TRO.2012.2234311}
}

@inproceedings{sabet2019rollocopter,
  author = {Sabet, Sahand and Agha-Mohammadi, Ali Akbar and Tagliabue, Andrea and
            Elliott, D. Sawyer and Nikravesh, Parviz E.},
  title = {{Rollocopter}: An Energy-Aware Hybrid Aerial-Ground Mobility
           for Extreme Terrains},
  booktitle = {2019 IEEE Aerospace Conference},
  pages = {1--8},
  year = {2019},
  doi = {10.1109/AERO.2019.8741685}
}

@article{ren2023spherical,
  title={Spherical robot: A novel robot for exploration in harsh unknown environments},
  author={Ren, Wei and Wang, You and Liu, Haoxiang and Jin, Song and Wang, Yixu and Liu, Yifan and Zhang, Ziang and Hu, Tao and Li, Guang},
  journal={IET Cyber-Systems and Robotics},
  volume={5},
  number={4},
  pages={e12099},
  year={2023},
  publisher={Wiley Online Library}
}

@article{shin2024fast,
  title={Fast ground-to-air transition with avian-inspired multifunctional legs},
  author={Shin, Won Dong and Phan, Hoang-Vu and Daley, Monica A and Ijspeert, Auke J and Floreano, Dario},
  journal={Nature},
  volume={636},
  number={8041},
  pages={86--91},
  year={2024},
  publisher={Nature Publishing Group UK London}
}

@inproceedings{liang2012amphihex,
  author = {Liang, Xu and Xu, Min and Xu, Lichao and Liu, Peng and
            Ren, Xiaoshuang and Kong, Ziwen and Yang, Jie and Zhang, Shiwu},
  title = {The {AmphiHex}: A Novel Amphibious Robot with Transformable
           Leg-Flipper Composite Propulsion Mechanism},
  booktitle = {2012 IEEE/RSJ International Conference on Intelligent Robots and Systems},
  pages = {3667--3672},
  year = {2012},
  doi = {10.1109/IROS.2012.6386238}
}

@inproceedings{kalantari2013hytaq,
  author = {Kalantari, Arash and Spenko, Matthew},
  title = {Design and Experimental Validation of {HyTAQ},
           a Hybrid Terrestrial and Aerial Quadrotor},
  booktitle = {2013 IEEE International Conference on Robotics and Automation},
  pages = {4445--4450},
  year = {2013},
  doi = {10.1109/ICRA.2013.6631208}
}

@article{daler2015bioinspired,
  author = {Daler, Ludovic and Mintchev, Stefano and
            Stefanini, Cesare and Floreano, Dario},
  title = {A Bioinspired Multi-Modal Flying and Walking Robot},
  journal = {Bioinspiration \& Biomimetics},
  volume = {10},
  number = {1},
  pages = {016005},
  year = {2015},
  doi = {10.1088/1748-3190/10/1/016005}
}

@article{sihite2023m4,
  author = {Sihite, Eric and Kalantari, Arash and Nemovi, Reza and
            Ramezani, Alireza and Gharib, Morteza},
  title = {Multi-Modal Mobility Morphobot ({M4}) with Appendage Repurposing
           for Locomotion Plasticity Enhancement},
  journal = {Nature Communications},
  volume = {14},
  pages = {3323},
  year = {2023},
  doi = {10.1038/s41467-023-39018-y}
}

@article{dudek2007aqua,
  author = {Dudek, Gregory and Giguere, Philippe and Prahacs, Chris and
            Saunderson, Shane and Sattar, Junaed and Torres-Mendez, Luz-Abril and
            Jenkin, Michael and German, Andrew and Hogue, Andrew and
            Ripsman, Arlene and Zacher, Jim and Milios, Evangelos and
            Liu, Hui and Zhang, Pifu and Buehler, Martin and Georgiades, Christina},
  title = {{AQUA}: An Amphibious Autonomous Robot},
  journal = {Computer},
  volume = {40},
  number = {1},
  pages = {46--53},
  year = {2007},
  doi = {10.1109/MC.2007.6}
}

@inproceedings{kim2026flashsac,
  author = {Kim, Donghu and Lee, Youngdo and Park, Minho and Kim, Kinam and
            Nahendra, I Made Aswin and Seno, Takuma and Min, Sehee and
            Palenicek, Daniel and Vogt, Florian and Kragic, Danica and
            Peters, Jan and Choo, Jaegul and Lee, Hojoon},
  title = {{FlashSAC}: Fast and Stable Off-Policy Reinforcement Learning
           for High-Dimensional Robot Control},
  booktitle = {Robotics: Science and Systems (RSS)},
  year = {2026},
  note = {arXiv:2604.04539}
}

@misc{zakka2026mjlab,
  author = {Zakka, Kevin and Liao, Qiayuan and Yi, Brent and Le Lay, Louis and
            Sreenath, Koushil and Abbeel, Pieter},
  title = {mjlab: A Lightweight Framework for {GPU}-Accelerated Robot Learning},
  year = {2026},
  eprint = {2601.22074},
  archivePrefix = {arXiv},
  primaryClass = {cs.RO},
  url = {https://arxiv.org/abs/2601.22074}
}

@inproceedings{todorov2012mujoco,
  author = {Todorov, Emanuel and Erez, Tom and Tassa, Yuval},
  title = {{MuJoCo}: A Physics Engine for Model-Based Control},
  booktitle = {IEEE/RSJ International Conference on Intelligent Robots and Systems (IROS)},
  pages = {5026--5033},
  year = {2012},
  doi = {10.1109/IROS.2012.6386109}
}
\end{document}